\documentclass{article}
\usepackage{spconf,amsmath,graphicx}

\RequirePackage{booktabs}

\usepackage{graphicx}
\usepackage{xcolor}
\usepackage{rotating}
\usepackage{multirow}
\usepackage{hhline}
\usepackage{url}
\usepackage{hyperref}

\title{Adversarial Example Detection by Classification for Deep Speech Recognition}

\name{Saeid Samizade$^{1,}$$^2$, Zheng-Hua Tan$^1$, Chao Shen$^2$ , Xiaohong Guan$^2$}
\address{
  $^1$Department of Electronic Systems, Aalborg University, Denmark\\
  $^2$MOE Key Lab for Intelligent Networks and Network Security, Xi’an Jiaotong University, China.\\
}

\begin{document}
\ninept

\maketitle
\begin{abstract}
 Machine Learning systems are vulnerable to adversarial attacks and will highly likely produce incorrect outputs under these attacks. There are white-box and black-box attacks regarding to adversary’s access level to the victim learning algorithm. 
 To defend the learning systems from these attacks, existing methods in the speech domain focus on modifying input signals and testing the behaviours of speech recognizers. We, however, formulate the defense as a classification problem and present a strategy for systematically generating adversarial example datasets: one for white-box attacks and one for black-box attacks, containing both adversarial and normal examples. The white-box attack is a gradient-based method on Baidu DeepSpeech with the Mozilla Common Voice database while the black-box attack is a gradient-free method on a deep model-based keyword spotting system with the Google Speech Command dataset.
 The generated datasets are used to train a proposed Convolutional Neural Network (CNN), together with cepstral features, to detect adversarial examples.  
  Experimental results show that, it is possible to accurately distinct between adversarial and normal examples for known attacks, in both single-condition and multi-condition training settings, while the performance degrades dramatically for unknown attacks. The adversarial datasets and the source code are made publicly available.
\end{abstract}

\section{Introduction}
\label{sec:intro}

Recent investigations in machine learning have demonstrated that machine learning systems are vulnerable against designed inputs known as adversarial examples \cite{Szegedy,ian}, a fundamental type of adversarial learning attack called test-time evasion \cite{miller2019adversarial}. Earlier adversarial attacks were applied on  machine learning models of image domain \cite{Liu,Carlini2017,ead,dongsu,MoosaviDezfooli2016DeepFoolAS} and then these attacking methods have been spread out onto other domains, e.g. speech signals \cite{carlinimain,Alzantot2017DidYH,Ciss2017HoudiniFD,kr,yuan}. The adversary adds a very small optimized perturbation, which is not detectable by human, to a legitimate input and generates an adversarial example that results the learning model to return a wrong output. It is therefore important to be able to detect adversarial attacks and subsequently for example prevent passing on the attacking signals to the learning model, a speech recognizer in the context of this work, so as to avoid outputting wrong results.

In general, there are two main categories of approaches in adversarial attacks as \emph{targeted'} and \emph{'non-targeted'} attacks. In non-targeted attacks the adversary aims to make the learning model return a wrong output and being wrong is enough. In targeted attacks the adversary aims to make the learning model return a particular output that is wrong and different from the expected output. Generating targeted examples is generally more difficult. This research focuses on targeted type of attacks on speech signal inputs for speech to text tasks. It is noted that adversarial examples in this context differ from data examples generated through generative adversarial networks \cite{goodfellow2014generative, michelsanti2017conditional} where examples are generated via adversarial training for the purpose of e.g. data augmentation and enhancement.

The adversary's level of access to the victim learning model categorizes attacking methods in two different types: 1) white-box where the adversary has full access to the layers and parameters of the victim learning model, and 2) black-box where the adversary has no access to these. The state-of-the-art white-box attack in the speech domain is Carlini \& Wagner (C\&W) method \cite{carlinimain} that is a gradient-based method using iterative optimization and has achieved 100\% success rate in their experiment. The similarity between normal and corresponding adversarial examples is 99\% when the victim model is Baidu DeepSpeech \cite{Battenberg2017ExploringNT} and the dataset is Mozilla Common Voice dataset \cite{Commonvoice}. As a black-box attack, Alzantot method \cite{Alzantot2017DidYH} is a gradient-free method using genetic algorithm optimization and has reported 87\% overall success rate. The similarity between normal and corresponding generated adversarial examples is 85\% when the victim model is Speech Command classification algorithm \cite{speechclass} and the dataset is Google Speech Command dataset \cite{Warden2018SpeechCA}.

Due to high success rates of adversarial attacks and their high similarity to normal examples, distinction of adversarial and normal examples is a highly motivated task. Research on adversarial attack detection or characterization has mostly focused on image domain \cite{Song2018PixelDefendLG,Xu2018FeatureSD}. One latest and comprehensive work in speech domain \cite{Yang2018CharacterizingAA} is to use temporal dependencies to characterize  adversarial examples. Defense methods, e.g. signal  transformation \cite{Yang2018CharacterizingAA} and obfuscated gradients \cite{Athalye2018ObfuscatedGG}, have also been investigated, but they provide rather limited robustness improvement in face of advanced attacks. Audio preprocessing methods and their ensemble for defense against black-box attacks are studied in \cite{rajaratnam2018isolated}. In \cite{rajaratnam2018noise}, noise flooding is applied to signals for defensing against black-box examples. All these methods are concerned with modifying input signals and testing the behaviours of the recognition model and have moderate success. There is a lack of studying dedicated systems for detecting both white-box and black-box adversarial examples in audio and speech domain.
This motivates us to formulate the defense against speech adversarial attacks as a classification problem, design a strategy for generating adversarial example datasets covering both black-box and white-box attacks and propose a CNN-based detection system. The created adversarial example datasets and the source code for adversarial example detection are made publicly available \footnote{\href{http://kom.aau.dk/~zt/online/adversarial_examples}{\scriptsize{\url{http://kom.aau.dk/~zt/online/adversarial\_examples}}}\par}. 

In generating adversarial examples, we take into account the length of source speech signals, the proportion of speech and non-speech in a signal,  and the length of targeted sentences and we consider both white-box and black-box attacks. For feature extraction, we apply cepstral features. 
For the detection model, we propose a CNN structure with small kernels in order to detect small perturbations in adversarial examples. To our knowledge, this is the first practical investigation on adversarial attack detection in speech recognition tasks. 

\section{Adversarial attack algorithms and dataset generation}

This section introduces two state-of-the-art attacking methods, one for white-box and another for black-box, and describes how we systematically generate datasets for detection of adversarial examples. 

\subsection{Attacking methods}


One of the most successful white-box attacking methods is C\&W \cite{carlinimain}. 
This method uses Connectionist Temporal Classification (CTC) loss function \cite{Graves:2006:CTC:1143844.1143891} for perturbation optimization. Like all white-box attacks, the adversary has full access to all layers and parameters of the speech recognition model and can use all gradients to minimize the perturbation while maximizing the success rate. Equation 1 shows the optimization of perturbation in the C\&W method:

\begin{equation}
\begin{split}
\text{minimize}_\delta \parallel \delta \parallel ^2 _2 + c.l(x+\delta, t)\\
\text{such that } dB_x(\delta) < \tau
\end{split}
\end{equation}

\noindent where $\delta$ is the added perturbation, $x$ is the legitimate speech example, $t$ is the desired target by adversary, $l(.)$ is the CTC loss function and the value $c$ is being updated to make a balance between changing the example $x$ to an adversarial example and keeping it close to the original normal example. The parameter $\tau$ is a threshold to indicate that the scale of perturbation $\delta$ should not be more than $\tau$ in order to keep the perturbation very low.

As an example for black-box attacking methods, we choose the Alzantot \cite{Alzantot2017DidYH} attacking method. This is a gradient-free method that uses a genetic algorithm. The method  has access only to the input and output of the victim speech recognition system. The difference between normal and adversarial examples for both attacking methods is illustrated in Fig. 1 using three spectrograms. The first one (a) belongs to a normal command example chosen from the Google Speech Command dataset. The original sentence of the sample is "yes". Spectrograms (b) and (c) are the generated adversarial examples with C\&W method and Alzantot method, respectively, using the same target word "right".

\begin{figure}[th]
  \centering
  \includegraphics[width=\linewidth]{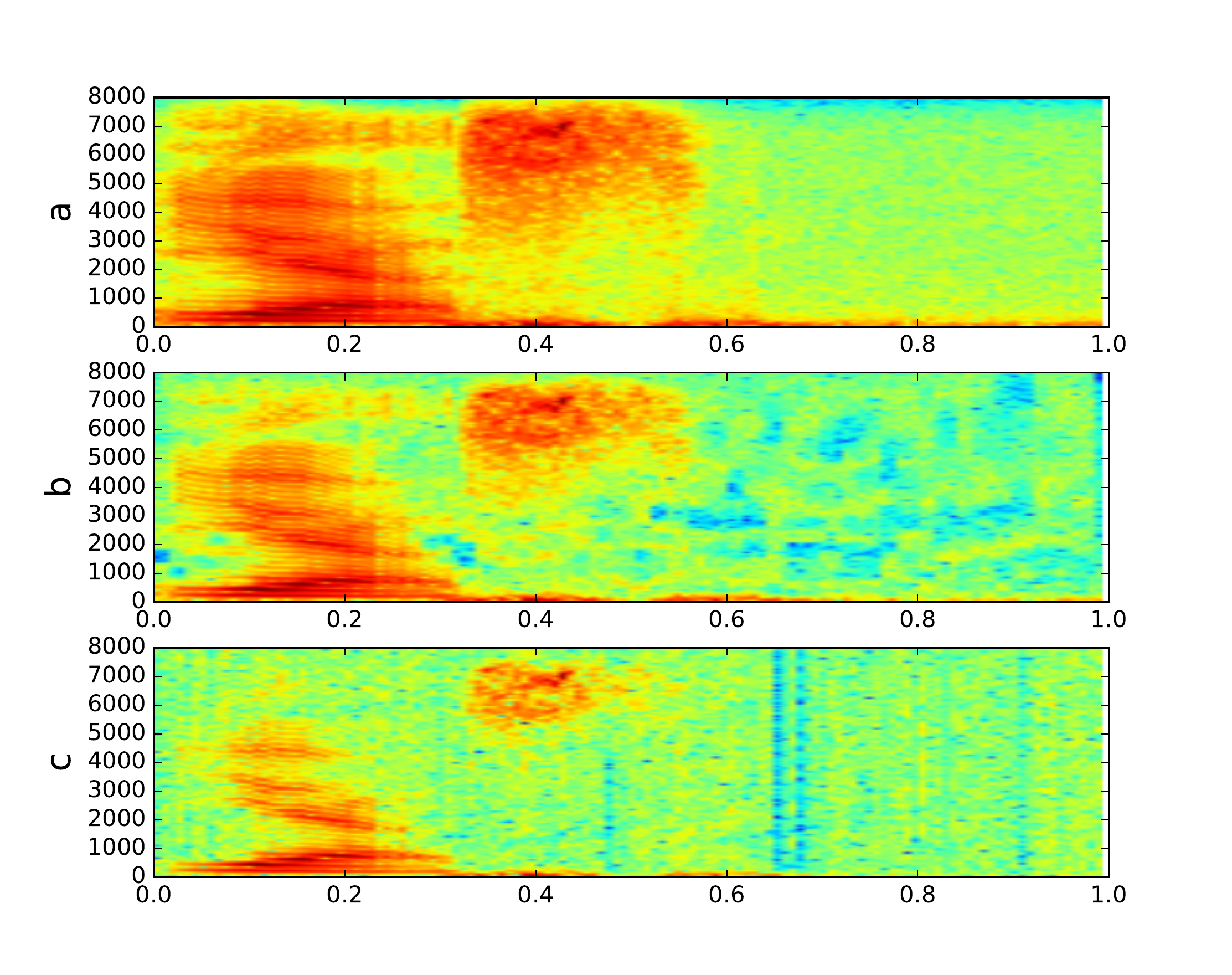}
  \caption{Spectrograms for normal example (a), white-box corresponding adversarial example (b) and black-box corresponding adversarial example (c). The original utterance of normal example is "yes" and the target for both adversarial examples is "right".}
  \label{fig:speech_production}
\end{figure}

Figure 1 shows that the C\&W method makes less changes in the speech region of original signal. The patterns of perturbation in the non-speech region are different in (b) and (c). When we listen to examples (b) and (c), we can clearly hear "yes" but low noise is present in the background. 

\subsection{Dataset generation}

Based on the two attacking methods introduced above, we design and generate two separated datasets: \emph{A} for white-box attacks and \emph{B} for black-box attacks. For white-box attacks, the C\&W method is chosen to attack  Baidu DeepSpeech \cite{Battenberg2017ExploringNT} as the victim learning model using  Mozilla Common Voice dataset \cite{Commonvoice}. As speech signals in  Mozilla Common Voice dataset have different lengths, we need choose from the huge dataset a subset of signals of various lengths in a principled way. According to our preliminary tests, there are three possibilities regarding to the length of original signals and the length of the target as follows: 1) Length of utterance is longer than length of target, 2) length of utterance is equal to length of target, and 3) length of utterance is shorter than length of target. 

Therefore, we categorize speech signals in the dataset to three different classes in the term of signal length: 
\begin{itemize}
\item \textbf{Short signals:} Audio files with the length of 1-2 seconds.
\item \textbf{Medium signals:} Audio files with the length of 3-4 seconds.
\item \textbf{Long signals:} Audio files with the length of 6-7 seconds.
\end{itemize}

The target of attack is other side of our equation. We thus also categorize our attacking targets to the same classes according to the common sentence lengths in the dataset and in this work compose the following targets:

\begin{itemize}
\item \textbf{Short target:} \emph{"Open all doors".}
\item \textbf{Medium target:}  \emph{"Switch off wifi connection".}
\item \textbf{Long target:}  \emph{"I need a reservation for sixteen people at the seafood restaurant down the street".}
\end{itemize}

Another consideration about choosing audio files is to ensure enough percent of speech region in the signal. There are some samples of minutes long but containing no speech and they are not good candidates for our dataset. We have chosen only audio files that contain more than 68\% of speech using the open-source robust Voice Activity Detection (rVAD) algorithm \cite{Tan2010LowComplexityVF}.

For each of the three categories of normal examples (short, medium and long), we chose 100 examples to be attacked. Each example was attacked by using the three targets (short, medium and long).
Consequently, we generated $100 \times 3 \times 3$ (i.e. 900) adversarial examples.  To have a balanced dataset without repeated samples, we chose 900 normal examples (300 examples for each category) that are different from the previously chosen 100 normal ones, to represent normal examples. In the end, we have the dataset \emph{A} containing 900 adversarial examples and 900 normal examples.

The black-box  method uses mutual targeting on Google Speech Command dataset \cite{Warden2018SpeechCA}, from which 10 different commands have been chosen. For each command, we use 20 samples. The attacking algorithm generates adversarial examples for each command using all other 9 commands as the target, so called mutual targeting.
As a result, we have 
 $10 \times 20 \times 9$ (i.e. 1800) adversarial examples in our generated dataset. Afterwards, we add other 1800 normal examples (180 examples for each command) to the dataset. The dataset \emph{B} thus contains 1800 normal and 1800 adversarial examples.
Note that all samples in both Mozilla Common Voice  and Google Speech Command datasets were recorded with 16kHz sampling rate and adversarial examples were generated with the same rate.

In this work, we used a Linux machine with four \emph{Nvidia GeForce GTX TITAN} GPUs. We applied the default parameters of original attacking methods to generate adversarial examples for both black-box and white-box attacks. The average time to generate a black-box adversarial example for a 1 second audio file is approximately 40 seconds and the average time to generate a white-box adversarial example for the same audio and the same target is around 4 minutes.

\section{Detection Method}

In this section we present the speech feature used in this work followed by our detection method.

\subsection{Speech Feature}


The Mel-frequency Cepstral Coefficient (MFCC) feature \cite{davis1980comparison} that has shown good performance across many tasks is used in this work, while cepstral features with different filter banks \cite{Yu2017DNNFB} are of interest for further study. The process of MFCC feature extraction involves dividing the signal into small frames, applying Mel-frequency filterbanks onto the frames, and taking logarithmic compression and discrete cosine transform. The extracted MFCCs of a signal have $f \times t$ dimensions where $f$ indicates the number of coefficients and $t$ indicates the number of frames. In this work we have $f = 40$ as a fixed value for all signals and $t$ varies across speech samples depending on their length. In this work we use as the classifier (detection model) a CNN model, which is widely used for speech applications \cite{CNN, speechclass}, but input samples of different sizes present a problem for CNN. However, this can be solved by using zero padding for all samples, and the dimension $t$ was set to the maximum value of our audio samples that is 698 (the longest audio files in \emph{long} category of the dataset \emph{A}). In practice, there are various ways to make this computation more efficient, which is out of the scope. 

\subsection{CNN architecture}

Let us consider the input speech signal represented in two dimensions $\textbf{V} \in \Re ^ {t \times f}$. 
The correlation in both time and cepstral dimensions are to be considered in a sliding receptive field. Therefore, we have a convolution of our input $\bf{V}$ and the weight matrix $\bf{W} \in \Re ^ {(m \times r) \times n}$, where $m \leq t$ and $r \leq f$ show the area of receptive field in time and cepstral directions, respectively. Consequently, we have $n$ feature maps with the size of $s \times v$ where $s$ and $v$ indicate the dimensions of feature maps.  

The proposed CNN architecture for this work, as shown in Fig. 2, is a simple and typical one that  uses three convolutional layers with small receptive fields. The first and the second 2D convolutional layers have the size of 64 with a \emph{Relu} activation function while the third one has the size of 32 with a linear \emph{Selu} activation function. A max-pooling layer is performed after each convolutional layer. Next, there are a fully connected layer with the size of 128 and a binary output softmax layer  to predict if the input is normal or adversarial. Figure 2 shows the architecture of the proposed CNN.

\begin{figure}[th]
  \centering
  \includegraphics[width=\linewidth]{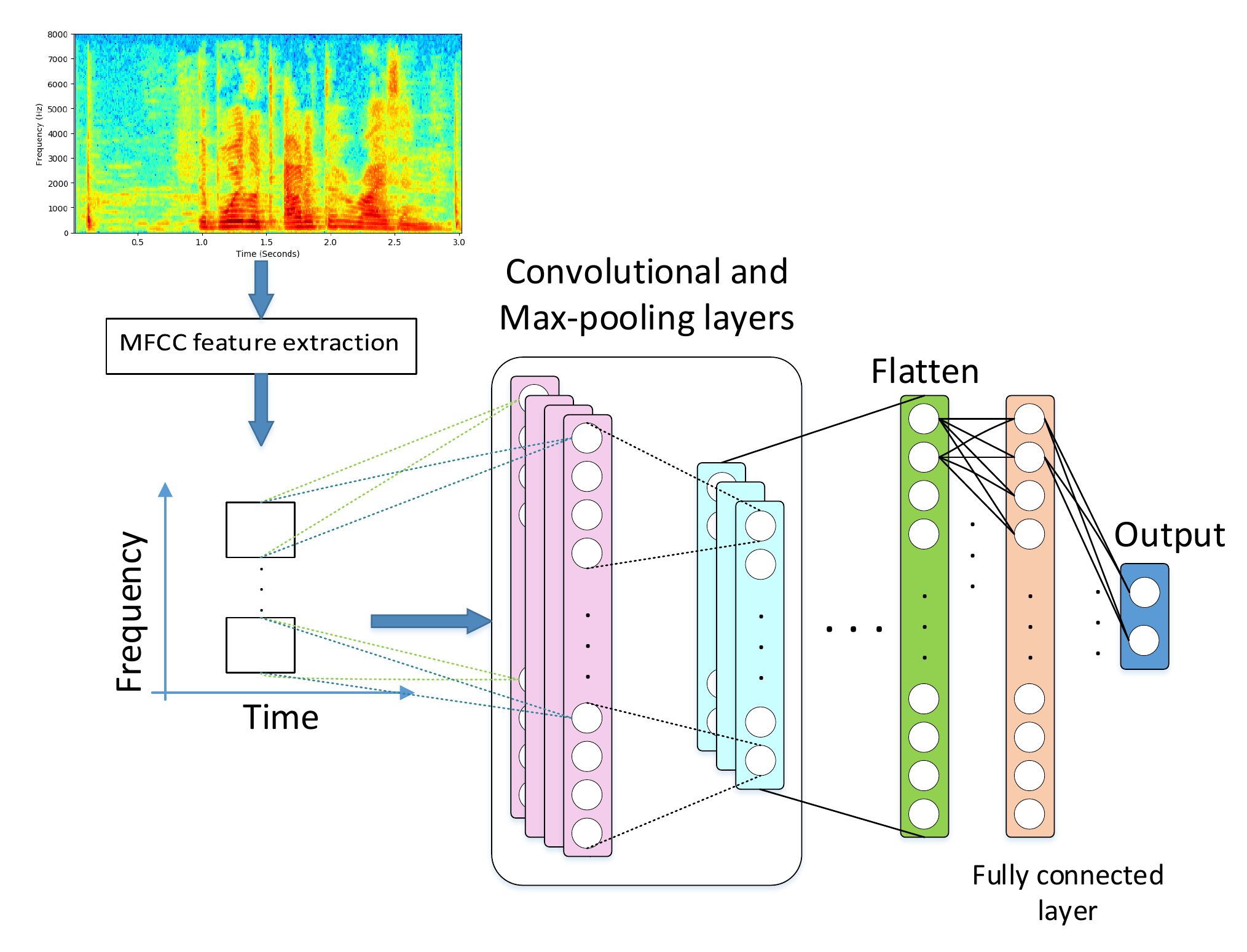}
  \caption{Architecture of the proposed CNN model.}
  \label{fig:speech_production}
\end{figure}

The CNN architecture is designed to be very sensitive to small changes in both time and cepstral directions. Consequently, it can recognize small changes with special patterns. We expect that this model performs well to detect adversarial perturbations. Detailed architecture of the proposed CNN model is presented in Table \ref{tab1}.
\vspace{-6pt}

\begin{table}[th]
  \caption{Detailed architecture of the proposed CNN model}
  \label{tab1}
  \centering
  \begin{tabular}{ l l l l }
    \toprule
    \textbf{type} & \textbf{size} & \textbf{field} & \textbf{activation}\\
    \midrule
      Convolutional 2D & 64 & kernel size=(2, 2)  & Relu\\
      MaxPooling 2D &  & pool size=(1, 3)  & \\
      Convolutional 2D & 64 & kernel size=(2, 2)  & Relu\\
      MaxPooling 2D &  & pool size=(1, 1)  & \\
      Convolutional 2D & 32 & kernel size=(2, 2)  & Linear\\
      MaxPooling 2D &  &  & \\
      Flatten &  &  & \\
      Fully connected & 128  &  & Relu \\
      Fully connected & 2 &  & Softmax \\
    
    \bottomrule
  \end{tabular}
\end{table}

\section{Experiments}

This section presents experimental settings, results and discussions.

\subsection{Experimental settings}

The experiments were conducted based on the two generated datasets \emph{A} (white-box) and \emph{B} (black-box). Six different experimental scenarios are designed as shown in the first three columns of Table \ref{tab2}. The goal is to test the detection performance of a narrowly trained system (only one type of attack) for both known and unknown attacks and the performance of a broadly trained system (both types of attack).  We expect to have high accuracy in the situations of matched training and testing 
while the performance for unknown attacks is expected to be low. Multi-condition training with \emph{A} and \emph{B} 
is expected to boost the performance of testing on both types of attacks with one system. 


\vspace{-6pt}

\begin{table}[th]
  \caption{The scenarios and accuracy results of testing data with  95\%  confidence interval for our experiments}
  \label{tab2}
  \centering
  \begin{tabular}{ l l l l l }
    \toprule
    \textbf{ID} & \textbf{Training} & \textbf{Testing} & \textbf{Accuracy \%} \\
    \midrule
    1                       & \emph{A} & \emph{A}  & $99.31 \pm 0.64$     \\
    2                       & \emph{A} & \emph{B}  & $82.07 \pm 1.52$     \\
    3                       & \emph{B} & \emph{A}   & $48.76 \pm 1.93$      \\
    4                       & \emph{B} & \emph{B}   & $98.42 \pm 0.86$       \\
    5                       & \emph{A, B} & \emph{A}  & $95.53 \pm 1.78$    \\
    6                       & \emph{A, B} & \emph{B} & $96.41 \pm 1.05$       \\
    \bottomrule
  \end{tabular}
\end{table}

\subsection{Training}

To evaluate our CNN detection model, we have separated each dataset (\emph{A} and \emph{B}) into \emph{Training} and \emph{Testing} subsets, with 75\% for training and 25\% for testing. It was also ensured that the adversarial examples in the training subset were generated from source examples that are different from those source examples used for generating testing adversarial examples. All training epochs are 100, which is enough for all models to converge. 


\subsection{Results and analysis}

To have more reliable results, we ran all experiments under the same conditions with the same configurations. Each experiment was run 10 times. The last column of Table \ref{tab2} shows overall accuracy numbers with 95\% confidence interval for all scenarios for testing data. The accuracy numbers of matched training and testing conditions (1 and 4) are more than 98\% and those of multi-condition training (5 and 6) are more than 96\%. 
These results show that the CNN model can learn adversarial perturbation very well although there exists noise in some normal speech examples. The CNN model can detect white-box examples better than black-box examples, which is slightly surprising since the overall difference between normal and adversarial examples is 99\%  for white-box  and it is 85\% for black-box. This result can be explained by looking at the adversarial perturbations generated by each method. The adversarial perturbation in white-box is very small but has special patterns (as shown in Fig. 1). However, the adversarial perturbation in black-box attack, although being larger, is more similar to normal noise in real world.

The results of mismatched training and testing conditions (2 and 3), as shown in Table \ref{tab2}, are very different. Accuracy for (Train A, Test B) is lower than matched conditions, but is still around 81\%. The accuracy of (Train B, Test A), however, is only a random guessing. This result indicates the perturbations of these two methods are quite different and dedicated training is required. 

The detailed results for the Experiment 1 (Train A, Test A) are shown in Table \ref{tab4}. It is obvious that the highest accuracy of detection happens when we have maximum difference between the length of source signal and the length of attacking target. Furthermore, when these two lengths are more equal, we have lower accuracy.

\vspace{-6pt}

\begin{table}[th]
\caption{Detailed accuracy results of testing data in percentage  for experiment (Train A, Test A) }
\label{tab4}
\centering
\begin{tabular}{ |l|c|c|c|c| } 
 \multicolumn{4}{r}{Targets}\\
 \hline
 & & Short & Medium & Long \\ 
 \hline

 \multirow{3}{*}{\begin{turn}{90}{Length}\end{turn}} & Short & 98.7 & 99.7 & 100 \\ 

 & Medium & 99.6 & 98.4 & 99.6 \\ 

 & Long & 100 & 99.2 & 98.4 \\ 
 \hline
\end{tabular}
\end{table}

Figure 3 illustrates the detailed results for Experiment 4 (Train B, Test B). It shows that when we have more similarity between source labels and target sentences in the term of phonemes, the accuracy of detection is lower.

\vspace{-6pt}

\begin{figure}[th]
  \centering
  \includegraphics[width=\linewidth]{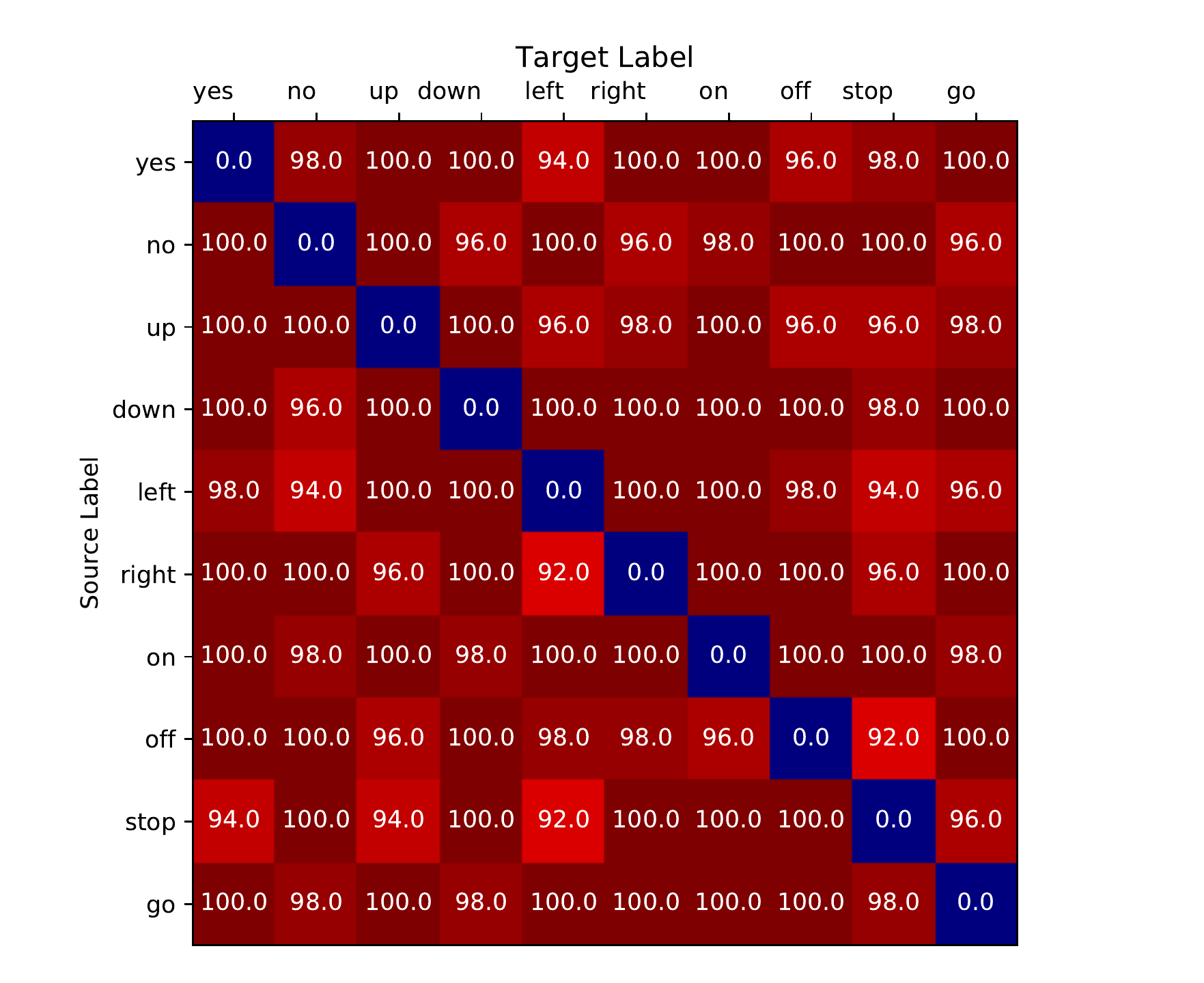}
  \caption{Detailed accuracy results of testing data in percentage for experiment (Train B, Test B).}
  \label{fig:speech_production}
\end{figure}

One further experiment was conducted to evaluate the  performance on unknown target sentences. Table \ref{tab5} shows the results of testing adversarial examples with unknown targets and lengths using dataset \emph{A}. The high accuracy results indicate that the model is robust in detecting targets of unknown sentences and unknown lengths. 


\vspace{-6pt}

\begin{table}[th]
  \caption{The accuracy results of testing adversarial examples with unknown targets and unknown lengths}
  \label{tab5}
  \centering
  \begin{tabular}{ l c   }
    \toprule
    \textbf{experiment} & \textbf{Accuracy \%}\\
    \midrule
    Train Short \& Medium, Test Long    & $99.44$   \\
    Train Short \& Long, Test Medium    & $99.82$   \\
    Train Medium \& Long, Test Short    & $99.26$   \\
    \bottomrule
  \end{tabular}
\end{table}

\section{Conclusion}

We have presented a strategy to systematically generate two separated datasets of adversarial speech examples using state of the art attacking methods, one for white-box attacks and one for black-box attacks. Then, we proposed a CNN model for adversarial attack detection and evaluated the model with the generated datasets through matched, mismatched and multi-condition training and testing strategies. Experimental results demonstrate that it is feasible to train a learning model to accurately detect adversarial examples generated from known attacking methods while detecting unknown attacks deserves more attention. The work further provides insights about the behaviours of different training strategies. 

\section{Acknowledgement}
We wish to thank the authors of the Mozilla Common Voice
dataset and the Google Speech
Command dataset, which our datasets are built upon, for making their datasets publicly available. 

\bibliographystyle{IEEEbib}

\bibliography{mybib}


\end{document}